\RequirePackage{fix-cm}
\documentclass[twocolumn]{svjour3}          
\smartqed  
\usepackage{natbib}
\usepackage{graphicx}
\usepackage{epstopdf}
\usepackage{amsmath}
\usepackage{amssymb}
\usepackage{fixltx2e}
\usepackage[hidelinks]{hyperref}
\usepackage{nicefrac}

\usepackage{color}

\begin{document}

\title{Automated Visual  Fin Identification \\ of  Individual Great White Sharks
}
\subtitle{}


\author{Benjamin Hughes and Tilo Burghardt
}


\institute{\textbf{Benjamin Hughes}\\ Dept of Computer Science,\ University of Bristol\\ Bristol BS8 1UB,\ United Kingdom\\ Now:\ Save Our Seas Foundation \\ Rue Philippe Plantamour 20,\ CH-1201 Geneva,\ Switzerland\\Email: \textit{ben@saveourseas.com}\\ \\ \textbf{Tilo Burghardt}\\ Dept of Computer Science,\ University of Bristol\\ Bristol BS8 1UB,\ United Kingdom\\ Email: \textit{tilo@cs.bris.ac.uk}}

\date{Received: date / Accepted: date}

\maketitle

\begin{abstract}
This paper discusses the    automated visual identification of individual great white sharks from dorsal fin imagery. We propose a computer vision photo ID system and  report recognition results over a database of thousands of unconstrained fin images. To the best of our knowledge this line of work establishes the first \textit{fully automated} contour-based visual ID system in the field of  animal biometrics. The approach put forward appreciates  shark fins  as  textureless, flexible and partially occluded objects with an individually  characteristic shape. In order to recover animal identities  from an image we first introduce an open contour stroke model, which extends multi-scale region segmentation to achieve robust fin detection. Secondly, we show that  combinatorial, scale-space selective fingerprinting can successfully encode fin individuality. We then measure the species-specific distribution of  visual individuality  along the fin contour via an embedding into a global    `fin space'. Exploiting this domain, we finally propose a non-linear model for individual animal recognition and combine all approaches into a fine-grained multi-instance  framework.  We provide a system  evaluation, compare results to prior work, and report   performance and properties in detail.      
\end{abstract}

\begin{figure*}
\includegraphics[width=1.0\textwidth]{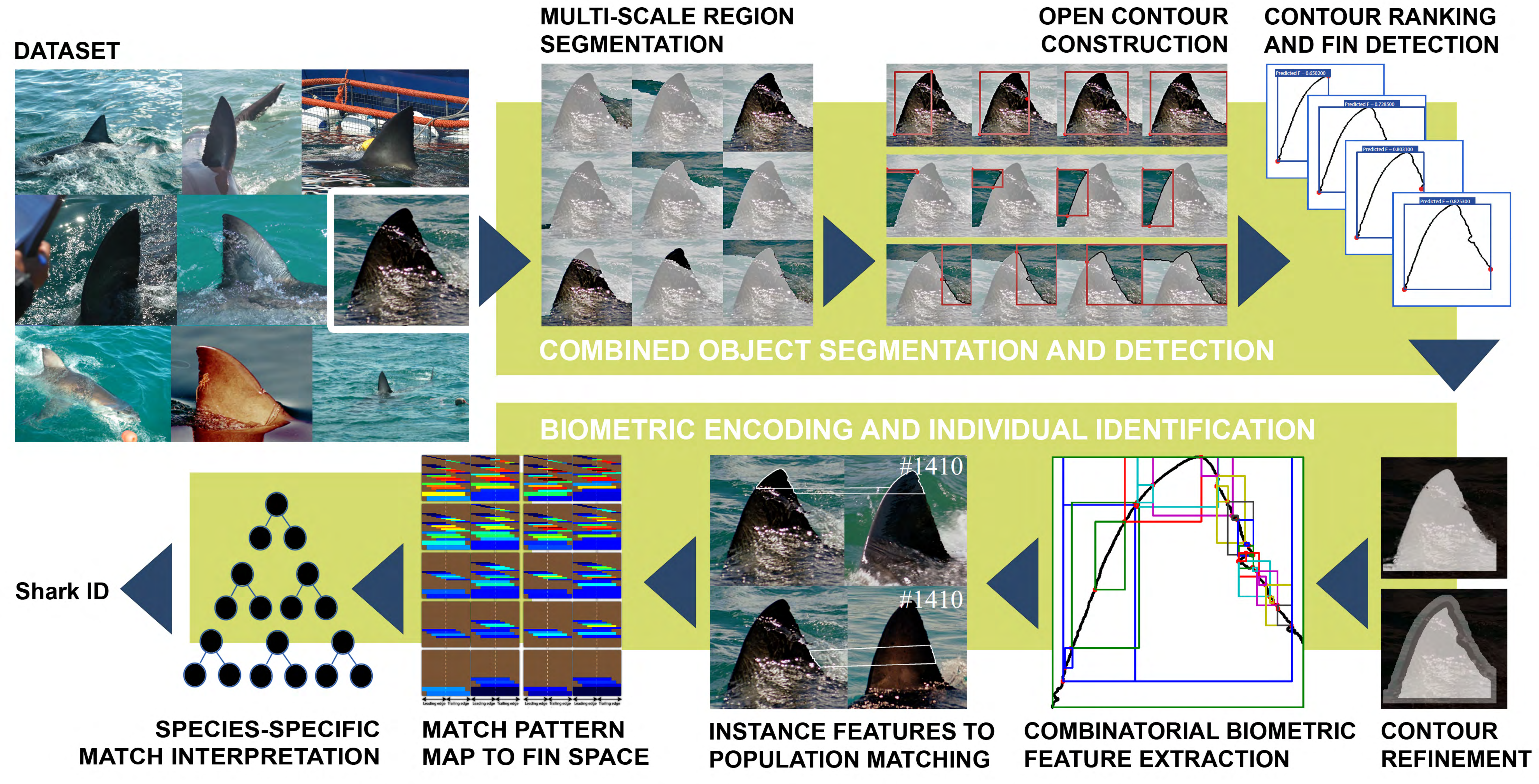}
\caption{ SYSTEM OVERVIEW: We perform a coarse and a fine-grained  recognition task. The first is to
simultaneously segment and detect shark fins, and the second is to recognise individuals. We begin by segmenting an image into an ultrametric contour map,
before partitioning boundaries into sets of open contours. We then train a random forest to rank contours and detect fin candidates based on normal information and opponentSIFT features. This forms the basis for computing individually distinctive  contour features, which are embedded into a species-specific  `fin space'. Shark identities are finally recovered by a non-linear, population-trained identification model that operates on this space.}
\label{fig:teaser}       
\end{figure*}

\section{Introduction}
\label{sec:intro}
Recognising individuals repeatedly over time is a basic requirement for field-based ecology and related life sciences \citep{marshall12}. In scenarios where photographic capture is feasible and animals are visually unique, biometric computer vision offers a non-invasive identification paradigm for handling this problem class efficiently \citep{Kuhl2013}. 
To act as an effective aid to biologists, these systems are required to operate reliably on large sets of unconstrained, natural imagery so as to facilitate adoption over widely available, manual or semi-manual identification systems \citep{stanley95,van07,ranguelova04,kelly01,speed07}. Further automation of identification pipelines for 2D biometric entities is currently subject to extensive  research activity~\citep{Duyck2014,Loos2013,Ravela2013}. Generally, fully automated approaches  require at least an integration of 1)~a robust fine-grained  detection framework to locate the animal or structure of interest in a natural image, and 2)~a biometric system to extract individuality-bearing features, normalise and match them \citep{Kuhl2013}. A recent example of such a system for   the identification  of  great apes \citep{Freytag2016,Loos2013} uses facial texture information to determine individuals. In fact, all fully automated systems so far rely on the presence of distinctive 2D colour and texture information for object detection as well as biometric analysis.    

In this paper we will focus on contour information of textureless objects as biometric entities instead. In specific, we propose a visual identification approach for great white shark fins as schematically outlined in Figure~\ref{fig:teaser}, one that extends work in~\cite{Hughes2015} and is applicable to unconstrained fin imagery. To the best of our knowledge this line of work establishes the first \textit{fully automated} \textit{contour-based visual ID system} in the field of  animal biometrics. It automates the
 pipeline from  natural image  to animal identity. We build on the fact that fin shape information has been used in the past manually to track individual great white sharks over prolonged periods of time \citep{anderson2011} or global space \citep{Bonfil2005}. Shark fin re-identification has  also been conducted semi-automati-cally to support research on the species \citep{towner13,chapple11,hillman03}. 

 We pose  the vision task of `shark fin identification' as a fine-grained, multi-instance classification problem for flexible, fairly textureless and possibly partly occluded object parts. 
`Fine-grained' in that each individual
fin, described by a characteristic shape and jagged trailing edge, is a subclass of the parent class great white shark fin. 
`Multi-instance' since the system must be able to assign  multiple semantic labels to an image,
 each label corresponding to an individual shark present. `Flexible' since fins  may bend, and `fairly textureless' since fins   lack distinctive
2D texture.
In line with work by \citet{arandjelovic11}, we will also refer to the  latter as `smooth'. We note that  some sharks carry fin pigmentation, yet not all do and its permanence is disputed \citep{robbins13}. Finally, fin detection  poses a part  recognition problem since region-based detection of the whole fin  would fail to tackle common scenarios: fins are often visually smoothly connected to the shark body whilst being partly occluded by the water line and white splash. Figure~\ref{fig:teaser} shows examples of the dataset~(top left) and outlines our solution pipeline based on contour information -- from image to individual shark ID. We will now review works closest related to the tasks of the recognition pipeline.

\vspace{-10pt}
\section{Related Work and Rationale}

\begin{figure*}
\vspace{-4pt}
\includegraphics[width=1.0\textwidth]{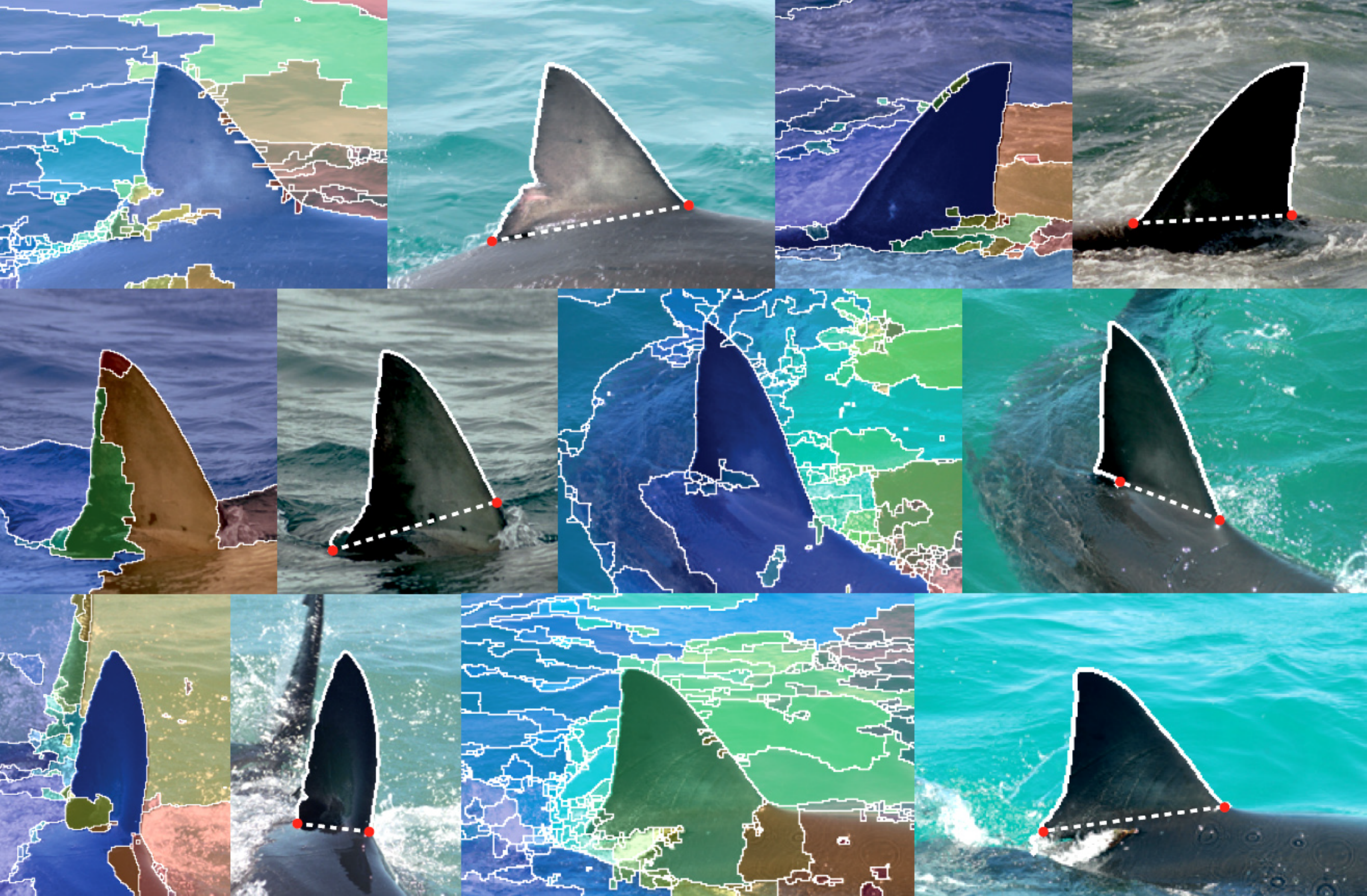}
\caption{FIN DETECTION AS OPEN\ CONTOUR STROKES: Multi-scale 2D~region-based segmentation algorithms~\cite{arbelaez14} on their own (left images show one level of the ultrametric map) regularly fail to detect the extent of fins due to visual ambiguities produced by shark body, water reflections or white splash. Thus, often no level of the underlying ultrametric contour map  captures  fin regions. We suggest combining properties of the 1D (open) contour segment shape with local 2D region structure in a contour stroke model to recognise the
fin section~(shown in solid white).\vspace{-2pt}}
\label{fig:teaser2}       
\end{figure*}

\begin{figure*}[ht]
\includegraphics[width=1.0\textwidth]{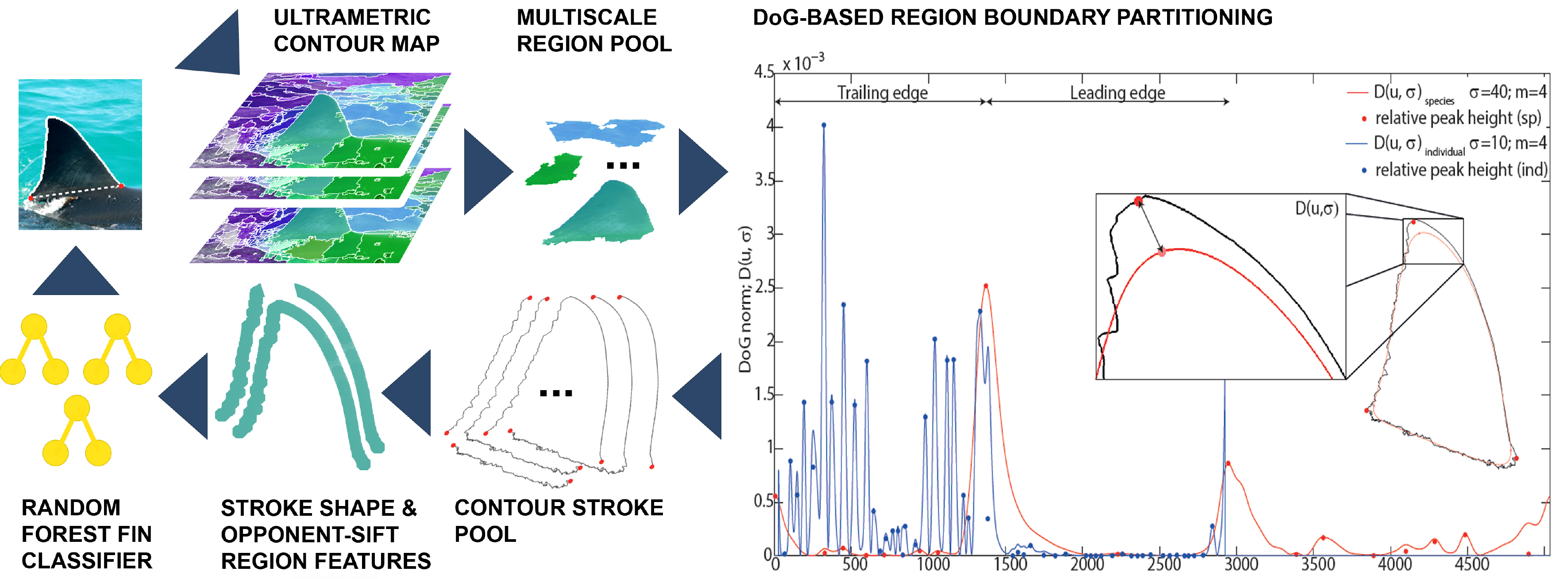}\vspace{-4pt}
\textcolor[rgb]{1,1,1}{..................................................................................................................................................}u
\caption{FIN DETECTION MODEL: Partitioning the (closed) 2D region structures  from across \textit{all} levels of the ultrametric contour map via DoG-generated keypoints (rightmost visualisation)  yields a pool of (open) contour strokes, whose normal-encoded shape and nearby opponentSIFT descriptors feed  into a random forest regressor to detect fin objects.}
\label{fig:teaser3}
\vspace{-8pt}       
\end{figure*}

\textbf{Smooth Object Detection.} Smooth object detection traditionally builds on utilising boundary and internal contour features, and  configurations thereof. Recent approaches \citep{arandjelovic11,arandjelovicO12} extend these base features by mechanisms for regionalising or  globalising   information, and infer object presence from learning configuration  classifiers. A prominent, recent example is Arandjelovic and Zisserman's~`Bag of Boundaries~(BoB)' approach \citep{arandjelovic11}, which employs multi-scale, semi-local shape-based boundary descriptors to regionalise BoB features and  predict object presence.

A related, more efficient boundary representation is proposed by \citet{arandjelovicO12}, which focusses
on a 1D semi-local description of boundary neighbourhoods around salient scale-space curvature maxima.
This description is based on a vector of boundary normals (Bag of Normals; BoN). However, experiments by \citet{arandjelovicO12} are run on images taken under controlled conditions \citep{geusebroek05}, whilst in our work, in common with \citet{arandjelovic11}, we have the goal of separating objects in natural images and against cluttered
backgrounds (see again Figure~\ref{fig:teaser}). 
\\ \ \\
\textbf{Fin Segmentation Considerations.} 
The biometric  problem at hand  requires an explicit, pixel-accurate encoding of the fin  boundary and sections thereof to  readily derive individually characteristic
descriptors. To achieve such   segmentation one could  utilise various approaches, including 1) a bottom-up
grouping process from which to generate object hypotheses for subsequent detection \citep{carreira10,li10,uijlings13,gu09},
or 2)~a~top-down sliding
window detector such as \citep{viola01,dalal05,felzenszwalb10} and then segment further detail,  or 3) combining the two simultaneously \citep{arbelaez12}. We  select the first option here since boundary encoding is intrinsic, and bottom-up, efficient and accurate object segmentation has recently become feasible. \cite{arbelaez14}
introduce a fast normalised cuts algorithm, which is used to globalise local edge responses
produced by the structured edge detector of \cite{dollar13}. 

However, since fins represent open contour structures (see Figure~\ref{fig:teaser2}) we require some form of (multi-scale) open contour generation, which is proposed, similar to~\cite{arandjelovicO12}, by  stipulating   key points along  closed contours of the ultrametric map as generated by~\cite{arbelaez14}. Our proposed contour stroke model
(see Section~\ref{sec:strokeModel}) then combines shape information along these open contour sections and  nearby regional information  to identify and segment  fin structures. Note that these are objects which are not present as segments at \textit{any} level of the ultrametric map. \\
\\
\textbf{Biometrics Context.} Most closely related within the animal biometrics literature are the computer-assisted
fin recognition systems; DARWIN \citep{stanley95,stewman06} and Finscan \citep{hillman03}. DARWIN has been applied to great white sharks \citep{towner13,chapple11} and bottlenose dolphins \citep{vanHoey13} while Finscan has been applied to false killer whales \citep{baird08}, bottlenose dolphins \citep{baird09} and great white sharks, among other species \citep{hillman03}. However both differ significantly from our work in that they rely on user interaction to detect and extract fin instances. Their fin descriptors are also sensitive to partial occlusions since they are represented by single, global reference encodings. Additionally, in the case of DARWIN, fin shape is encoded as 2D Cartesian coordinates, requiring the use of pairwise correspondence matching. By contrast, we introduce an occlusion robust vector representation of semi-local fin shape (see Section~\ref{sec:idEncoding}). As in \cite{crall13}, this allows  images of individuals to be held in~tree-based search structures, which facilitate identity discovery in sub-linear time.
\\ \ \\
\textbf{Paper Structure.} The paper covers six further  sections, which will detail the methodology and algorithms proposed, and report on application results and  discuss our approach in its wider context. In (\ref{sec:strokeModel}), in accordance with~\cite{Hughes2015}, a contour stroke model for fin detection is presented combining a partitioning of ultrametric contour maps with normal descriptors and dense local features. Then, expanding on previous work, in (\ref{sec:idEncoding}) and  (\ref{sec:LNBNN}) a dual biometric encoding scheme for fins and an associated LNBNN baseline identification approach are discussed.  In (\ref{sec:finspace}) we quantify species-specific visual individuality via a `fin space', and  in (\ref{sec:ID}) an improved non-linear identification framework that uses this space is shown and evaluated. Finally, in~(\ref{sec:conclusions}) we discuss the scope and conclusions of individually identifying great white sharks visually.

\vspace{-12pt}
\section{Contour Stroke Object Model}
\label{sec:strokeModel}

\begin{figure}[hbt]
\includegraphics[width=84mm]{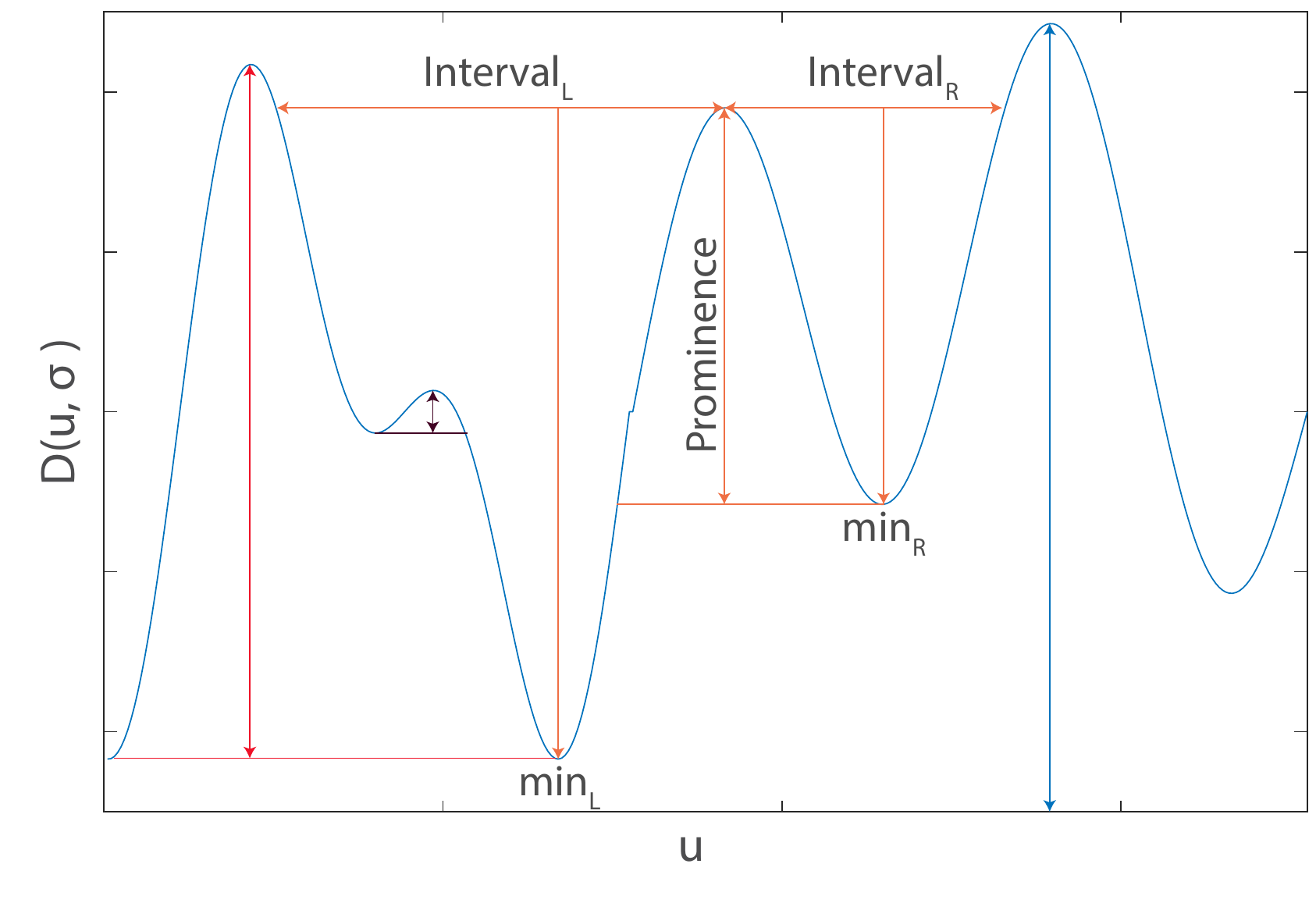}
\caption{NON-MAXIMUM SUPPRESSION: We utilise the  Matlab function `findpeaks' as a reference implementation for non-maximum suppression. That is, from a local maximum on $D(u,\sigma)$, the horizontal distance to $D(u,\sigma)$ is measured to define left and right intervals: min\textsubscript{L}=$\text{min}\textsubscript{interval\textsubscript{L}}(D(u,\sigma))$, min\textsubscript{R} is defined likewise. Subsequently, max(min\textsubscript{L},min\textsubscript{R}) is taken as a reference level. The prominence of each local maximum is then computed as the difference between the value of $D(u,\sigma)$ at the local maximum and the reference level. Low prominence peaks are suppressed. If either interval reaches the end of the signal, we set its minimum  to be zero. \vspace{-7pt}
}
\label{fig:nonmax}       
\end{figure}

\begin{figure}[t]
\vspace{-5pt}
\includegraphics[width=84mm]{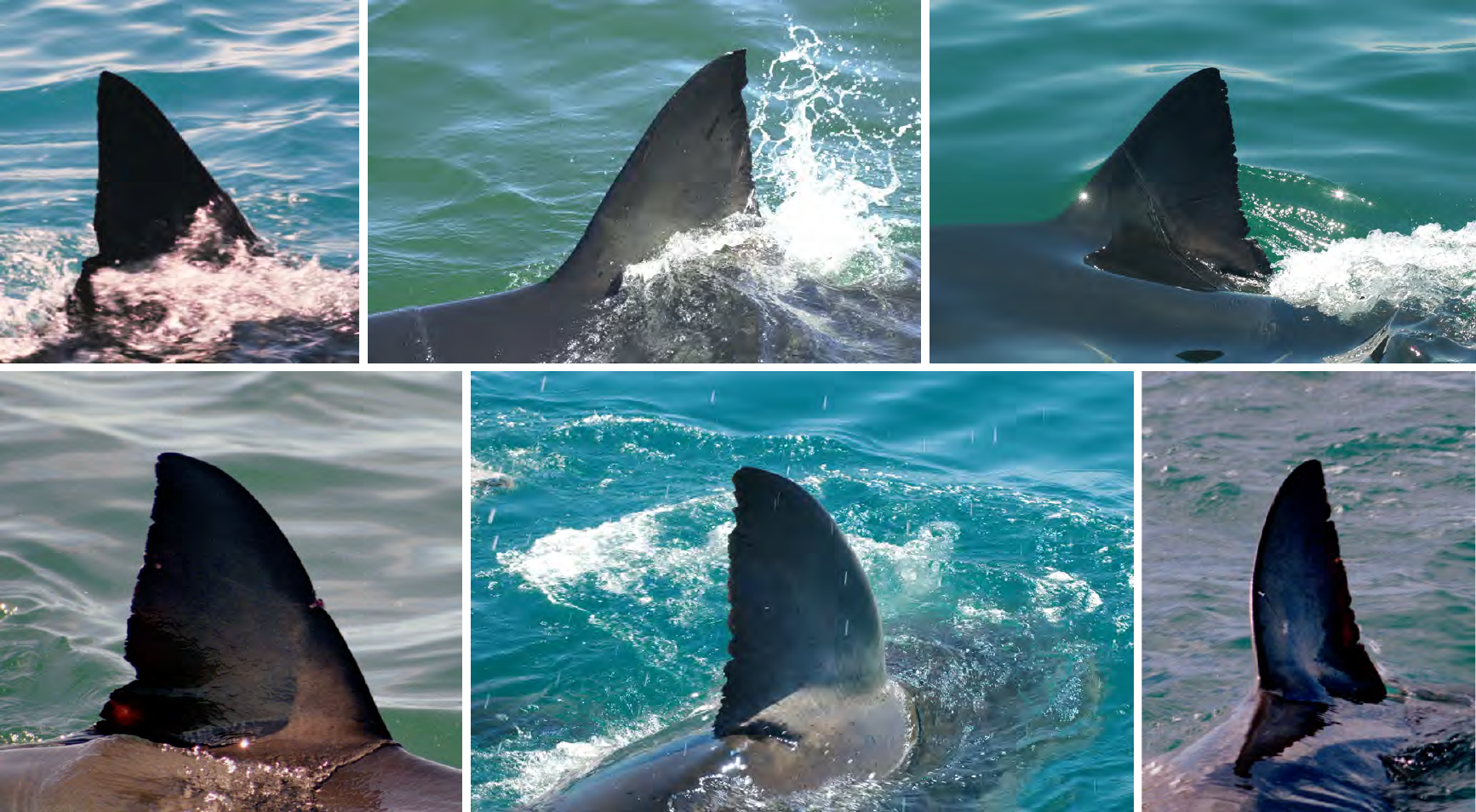}
\caption{HIGH AND LOWER VISIBILITY FIN IMAGES: The top row shows examples of lower visibility fin images where parts of the fin are occluded by water line and white splash. The bottom row shows high visibility fin images -- the entire  extent of the fin is visible.\vspace{-8pt}}
\label{fig:hiLowVis}       
\end{figure}

In this section we describe our contour stroke model for bottom-up fin detection.  It constructs fin candidates as subsections (or `strokes') of contours in partitioned ultrametric  maps and validates them by regression of associated stroke properties. The approach progresses in three stages: 1)~ we detect and group object boundaries at multiple scales into an ultrametric contour map, 2)~salient boundary locations are detected and used to partition region boundaries into  contour sections called strokes, 3)~strokes are classified into fin and background classes based on shape, encoded by normals, and local appearance encoded by opponentSIFT features \citep{van10}.
Figure~\ref{fig:teaser3} illustrates this fin detection approach in detail. \\ \ \\
\textbf{Stage 1:   Hierarchical Segmentation.} We use work by~\cite{arbelaez14} to generate a region hierarchy in the form of an ultrametric  map. This provides  sets of closed contours for any chosen level-threshold in the range $[0,1]$. Starting with the whole image, we descend the hierarchy to a pool of $200$ unique regions. Similar to~\cite{carreira10}, we then employ region rejection to remove areas too small to represent a fin, or too similar to another region\footnote{Any region with a boundary length of less than $70$ pixels is discarded, before the remainder are clustered into groups where all regions in a cluster have an overlap of $0.95$ or more. Within each cluster, we rank regions according to the level in the hierarchy at which they first appeared, retaining the top ranked region in each cluster. }.
 We subsequently rank  remaining regions, again by their location in the hierarchy, and retain the top $k$ regions, choosing $k=12$ empirically for the results given in this paper.
\\ \ \\
\textbf{Stage 2: Generating Fin Candidates.} In almost all cases, the segmentation produces at least one single region, within the set, that provides a high recall description of the fin's external boundary. However, in cases where the boundary between the fin and the body is visually smooth,  segmentation tends to group both in a single region  (see Figure~\ref{fig:teaser2}). The global appearance of such regions can vary dramatically, making 2D structures unsuitable targets for recognition. By contrast, locations along the 1D contour of regions provide discontinuities in curvature suitable for region sub-sectioning and thereby stroke generation. We detect boundary keypoints using the Difference of Gaussian (DoG) corner detector of \citet{zhang09}. Letting $C(u) = (x(u),y(u))$ represent a planar curve, the corner response function is given by the evolution difference of two Gaussian smoothed planar curves, measured using the distance $D(u,\sigma)$:\vspace{-5pt}
\begin{multline}
D(u,\sigma)=[DoG*x(u)]^2 + [DoG*y(u)]^2\\
 = [G(u,m\sigma)*x(u) - G(u,\sigma)*x(u)]^2+\\
 [G(u,m\sigma)*y(u) - G(u,\sigma)*y(u)]^2
 \label{eq:dog}
\end{multline}
where $G(u,\sigma)$ is a zero mean Gaussian function with standard deviation $\sigma$, and $m>0$ is a multiplication factor. Viewed as a bandpass filter, by varying $m$ and~$\sigma$, the operator can be tuned to different frequency components of contour shape. For keypoint detection (visualised rightmost in Figure~\ref{fig:teaser3}), we resample contours to $128$ pixels and compute $D$ using $\sigma=1$ and $m=4$ before ranking the local maxima of $D$ by their prominence (see Figure \ref{fig:nonmax}). This allows for the selection of the $n$ peaks with largest prominence suppressing other, locally non-maximal corner responses. Choosing small values of $\sigma$ ensures accurate keypoint localisation whilst a relatively large value of $m$ ensures that the $n$ largest maxima of $D$ correspond to globally salient locations. 
We then generate fin candidates as contour strokes by sampling the region contour between every permutation of keypoint pairs. This results in a pool of $ N_c=(n^2-n)k$ strokes per image without taking the two  encoding directions (clockwise and anticlockwise) into account. We set $n$ by assessing the \textit{achievable quality} (the quality of the best candidate as selected by an oracle) of the candidate pool with respect to the number of candidates. We denote this fin-like quality of stroke candidates by $F^g\textsubscript{inst}$. 
Evaluated with respect to a human-labelled ground truth contour, we use the standard $F$-measure for evaluating contour detections based on bipartite matching of boundary pixels \citep{martin04}. We observe that average achievable quality does not increase beyond $n=7$ given the described DoG parameterisation and therefore use this value to define~$N_c$. The result is that, on average, we obtain $504$ candidates per image, with an average achievable quality of $F^g\textsubscript{inst}=0.97$ measured against human-labelled ground truth contours for $240$ randomly selected images. By means of comparison, the average  quality of the pool of $k=12$ closed region contours is $F^g\textsubscript{inst}=0.75$. 
\ \\ \\
\textbf{Stage 3:  Fin Candidate Scoring.} For training and testing the candidate classifier, $240$ high visibility~(H) images, where the whole fin could clearly be seen \textit{above} the waterline, are selected at random and then randomly assigned to either a training or validation set, each containing $120$ images. In addition, we perform validation using a second set of $165$ `lower' visibility (L) images where fins are partially occluded, again, selected at random. This will enable us to establish whether the trained model is representative given partial occlusion. Examples of each image type are shown in Figure~\ref{fig:hiLowVis}. 

Ground truth fin boundary locations are labelled by hand using a single, continuous contour,~$1$~pixel~in~width. 
Each contour section is described by a $180$-dimensional feature vector consisting of two components, contributing 2D and 1D distinctive information, respectively. 

The first is a bag of opponentSIFT \citep{van10} visual words (dictionary size $20$) computed at multiple scales (patch sizes $16,24,32,40$) centred at every pixel within a distance of $4$ pixels of the contour section. This descriptor is utilised to capture the local appearance of fin contours.
The second describes contour shape using a histogram of boundary normals consisting of $20$ spatial bins and $8$~orientation bins. Note that the opponentSIFT histogram is independent of encoding direction whilst the histogram of boundary normals is dependent on it\footnote{When training the histogram of boundary model, we flip images so the fin is facing the same way in each. For testing, we compute two feature vectors, one for each fin direction. We then obtain a predicted quality score for each direction and take the maximum over directions as the predicted quality for that stroke.}.  

\begin{figure}[]
\vspace{-5pt}
\includegraphics[width=80mm]{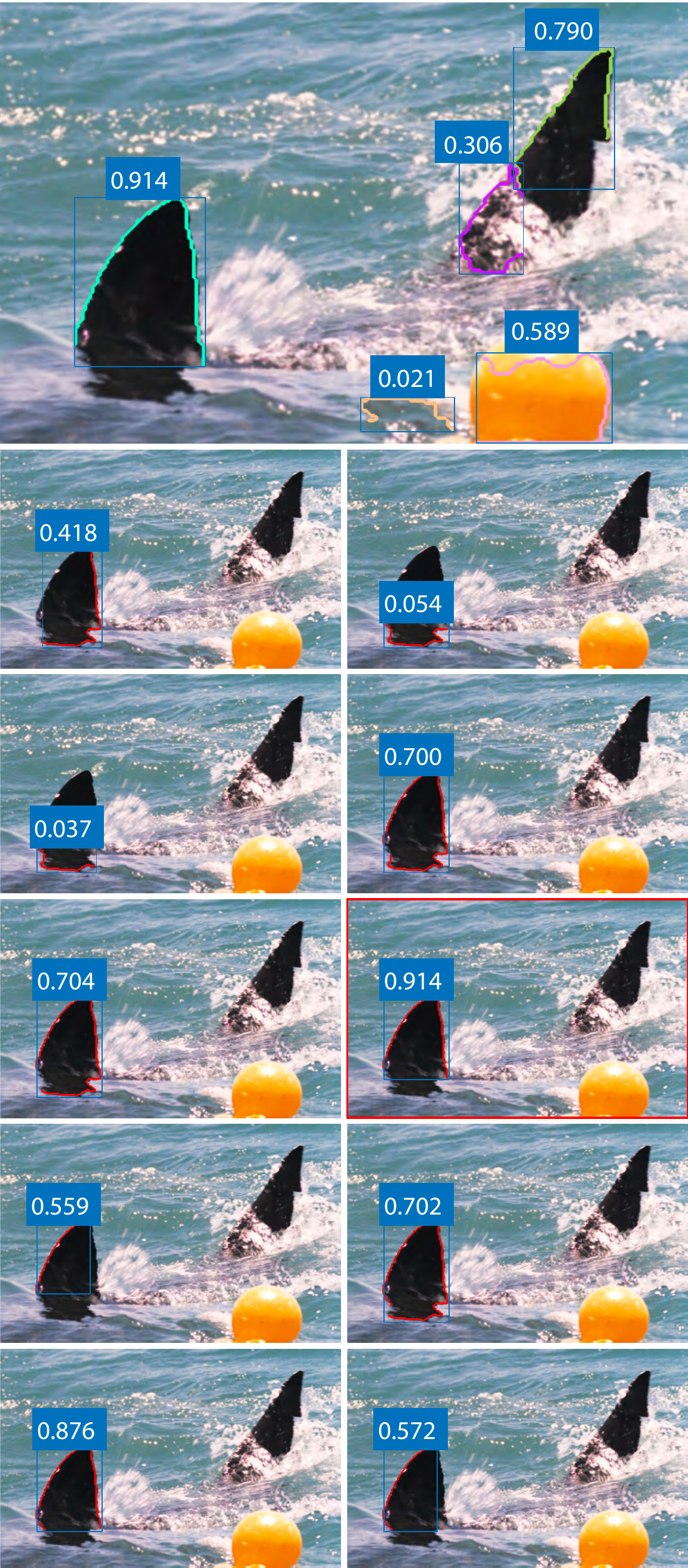}
\caption{ EXAMPLE FIN CANDIDATES AND  PREDICTED QUALITY ($F^p\textsubscript{inst}$). (Top)  Candidates and their scores after non-maximum suppression. (Other) Candidates  and scores from region around the fin before non-maximum suppression. The predictive ability of the model is reflected in the stroke quality predictions for strokes describing at least part of the fin. It is unsurprising that the model makes high quality-predictions for the caudal fin stroke. We also see that while higher scores are sometimes predicted for purely background objects, the scores predicted for these are typically not as high as those predicted for good quality strokes describing fins themselves.\vspace{-10pt} }
\label{fig:candidates}       
\end{figure}

In either case, the two components are $L_2$ normalised and concatenated to produce the final descriptor. A random forest regressor \citep{breiman01} is  trained to predict the quality of fin hypotheses where the quality of individual candidates is assessed using the $F$-measure as computed using the~BSDS contour detection evaluation framework \citep{martin04}. Following non-maximum suppression with a contour overlap threshold of $0.2$, a final classification is made by thresholding the predicted quality score.
Given an image, the resulting detector then produces a set of candidate detections, each with a predicted quality score $F^p\textsubscript{inst}$.
Figure~\ref{fig:candidates} illustrates
example candidates together with their scores.
\\
\textbf{Measuring Detection Performance.}  We use 1) average precision~(AP$^{t}_{det}$), the area under the precision-recall~(PR) curve for a given threshold $t$, and 2) volume under PR surface~(AP\textsuperscript{vol}) as evaluation metrics.
\begin{figure*}[hbt]
\vspace{-4pt}
\includegraphics[width=1.0\textwidth]{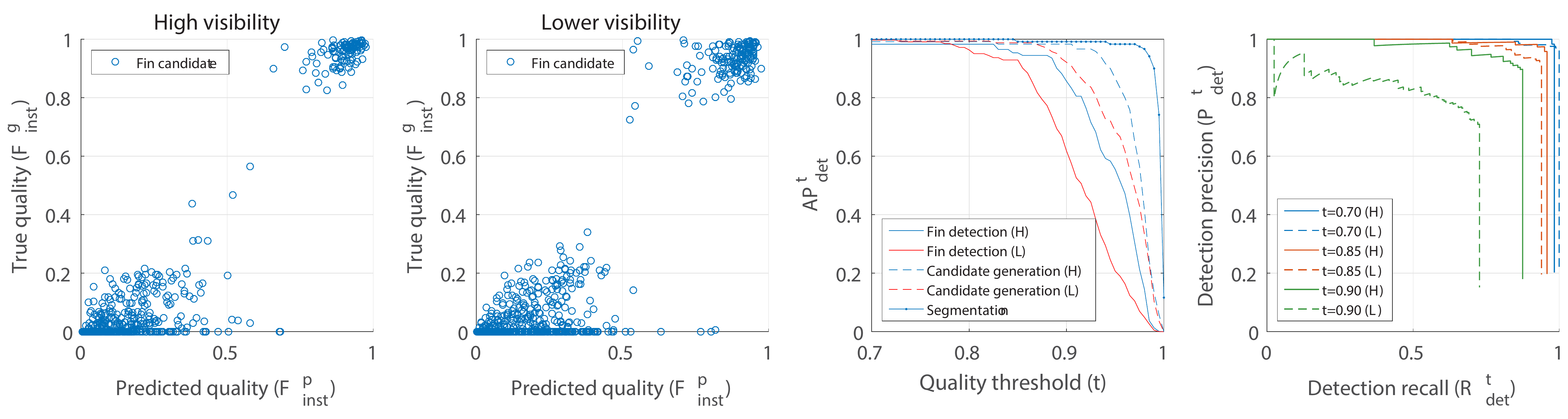}\\
\textcolor[rgb]{1,1,1}{.........................}(A)\textcolor[rgb]{1,1,1}{.............................................}(B)
\textcolor[rgb]{1,1,1}{............................................}(C)
\textcolor[rgb]{1,1,1}{...........................................}(D)
\caption{FIN DETECTION RESULTS: (A,B) Scatter plots  show that the full fin detection model is able strongly to  predict, as captured by~$F^p\textsubscript{inst}$, the true quality of fin candidates  $F^g\textsubscript{inst}$ for both high and low visibility images.  (C) The plot summarises performance at different stages of fin detection. Note, that for the `segmentation' line, AP$^{t}_{det}$ is equivalent to the proportion of fins for which it is possible to obtain a stroke of quality $F^g\textsubscript{inst} \geq t$, given a machine generated segmentation. (D) The plot shows PR plots for both high and low visibility images at different thresholds.
}
\label{fig:finDetectResults}       
\end{figure*} In order to generalise AP$^{t}_{det}$, the AP\textsuperscript{vol}  measure was proposed by \cite{hariharan14} for simultaneous object detection and segmentation. It measures the volume under a PR surface traced out by PR curves generated for variable quality thresholds $t$, and thus avoids arbitrary threshold choices. It reflects both fin detection performance and the quality of candidates detected and, as noted by \cite{hariharan14}, has the attractive property that a value of $1$ indicates perfect candidate generation as well as  fin detection.

We base our fin detection evaluation on AP instead of a receiver operating characteristic (ROC) curve  based measure such as AUC-ROC, since the choice of precision over FPR increases evaluation sensitivity to changing numbers of false positives in the presence of large numbers of negative examples~\citep{davis06}. In addition, the choice of AP-based evaluation is in line, not only with \cite{hariharan14}, but also with the  methodology adopted in the object detection components of the ImageNet Large Scale Visual Recognition Challenge (ILSVRC) \citep{russakovsky15} and in the PASCAL Visual Object Challenge (PASCAL VOC) \citep{everingham10}, two standard benchmarks for visual object recognition.
\\ \ \\
\textbf{Fin Detection Results.} Results for fin candidate generation and detection are shown in Figure~\ref{fig:finDetectResults}, Table~\ref{tab:intermediateResults} and Table~\ref{tab:detectResults}.
  Scatter plots in Figure~\ref{fig:finDetectResults} for high and lower visibility images confirm that the model is able strongly to identify fins, and many high quality candidates are generated as shown by the large number of instances with high $F^g\textsubscript{inst}$ scores. The Pearson correlation coefficients between true and predicted quality scores are 0.95 and 0.93, respectively.

The plot of Figure~\ref{fig:finDetectResults}(C) summarises performance at different stages of fin detection. We note that for segmentation, a stroke of quality $F^g\textsubscript{inst}\geq 0.95$ is possible for almost all fin instances ($98.3\%$), with an average achievable quality, AP\textsuperscript{vol}, of 0.99.
Candidate generation also performs well. It can be seen that for almost all high visibility fins ($98.3\%$), a candidate of $F^g\textsubscript{inst}> 0.9$ is generated and $F^g\textsubscript{inst}> 0.85$ for $98.8\%$ of lower visibility fins. Across all thresholds and fin instances, average achievable qualities of 0.97 and 0.96 are seen respectively. 
Table~\ref{tab:intermediateResults} summarises these intermediate results. 

Finally, we show results for the whole pipeline in Figure~\ref{fig:finDetectResults}(C) and Table~\ref{tab:detectResults}, that of combined segmentation, candidate generation and candidate classification. Here we see that a candidate of quality $F^g\textsubscript{inst} \geq 0.83$ is generated and recognised (with AP$^{t}_{det} =0.98$) for almost all high visibility fins ($F^g\textsubscript{inst} \geq 0.78$ for lower visibility with AP$^{t}_{det} =0.99$), as indicated by AP$^{t}_{det}$ close to $1$ for these quality thresholds, with AP$^{t}_{det} =1$ only possible if both P$^{t}_{det}=1$ and R$^{t}_{det} =1$.

\begin{table}[b]
\caption{INTERMEDIATE RESULTS (AP$^{t}_{det}$)}
\label{tab:intermediateResults}       
\begin{tabular}{lllll}
\hline\noalign{\smallskip}
 & t=0.7  & t=0.85 & t=0.9 & AP\textsuperscript{vol}  \\
\noalign{\smallskip}\hline\noalign{\smallskip}
Segmentation & 1.0  & 0.99 & 0.99 & 0.99\\
Candidate gen. (H) & 0.99  & 0.98 & 0.98 & 0.97 \\
Candidate gen. (L) & 1.0  & 0.99 & 0.92 & 0.96 \\
\noalign{\smallskip}\hline
\end{tabular}\vspace{10pt}
\end{table}

\begin{table}[b]
\caption{FIN DETECTION  RESULTS (AP$^{t}_{det}$)}
\label{tab:detectResults}       
\begin{tabular}{lllll}
\hline\noalign{\smallskip}
Feature type & t=0.7  & t=0.85 & t=0.9 & AP\textsuperscript{vol}  \\
\noalign{\smallskip}\hline\noalign{\smallskip}
\textbf{High Visibility (H)} &  &  &  &  \\
OpponentSIFT & 0.99  & 0.85 & 0.73 & -\\
Normal  & 0.98  & 0.85 & 0.7 & - \\
Combined & 0.98  & 0.95 & 0.86 & 0.92 \\
\noalign{\smallskip}\hline\noalign{\smallskip}
\textbf{Lower Visibility (L)} &  &  &  &  \\
Combined & 1.0  & 0.93 & 0.62 & 0.89  \\
\noalign{\smallskip}\hline
\end{tabular}
\end{table}

To fully understand values of AP$^{t}_{det}<1$, we must consider detection precision and recall separately, as shown in Figure~\ref{fig:finDetectResults}(D). Here we show PR curves for selected quality thresholds of the complete detection pipeline. We see for example that for $t=0.85$, perfect precision (P$^{t}_{det} =1.0$) is achieved for about 63\% of both high and lower visibility fins (R$^{t}_{det} =0.63$), after which, false positives are introduced as shown by reduced precision. We also see that detection recall does not equal~$1$ for any value of precision, repeating the observation that a candidate of this quality is not generated for every fin.   Meanwhile, we see near perfect detection if we accept candidates with $F^g\textsubscript{inst} \geq 0.7$.

Finally, observing the summary of results in Table~\ref{tab:detectResults}, we see the effectiveness of the different features types for fin candidate classification. It can be seen that while both opponentSIFT and normal features enable good detection performance (say for $t=0.7$), a combination of the two is required to obtain good recognition of the highest quality candidates at $t=0.9$.
In summary, for almost all fin instances, a high quality candidate is generated and recognised with high precision, demonstrating the effectiveness of our contour stroke model for the task at hand.  
\ \\ \\

\begin{figure*}[ht]
\includegraphics{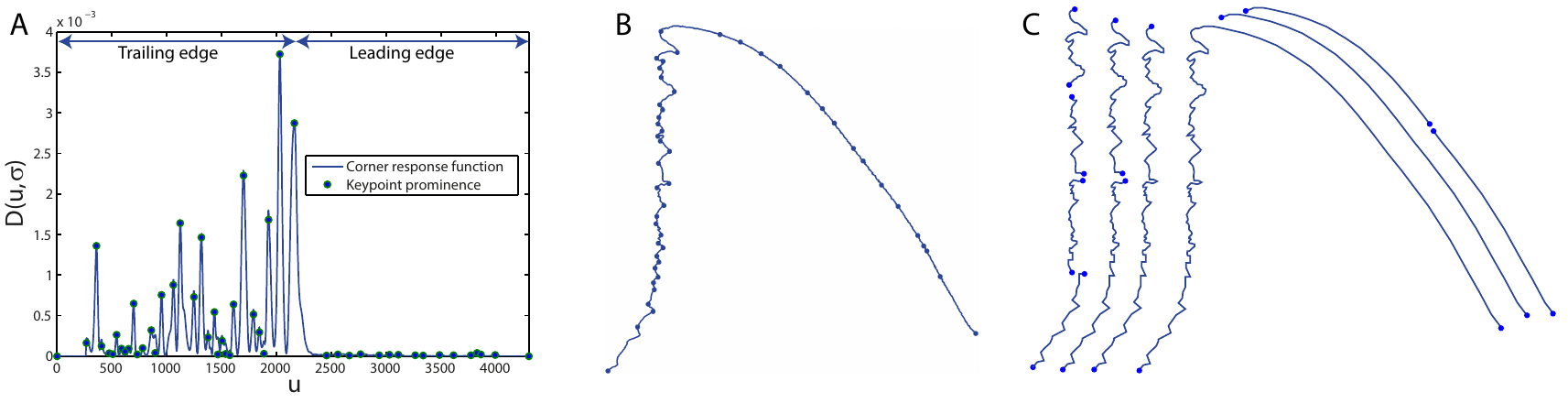}
\caption{COMBINATORIAL CONTOUR SAMPLING: (A) The DoG corner response function of a fin contour. (B) The $n=50$ most prominent maxima of $D$ are selected as keypoints. The detected keypoints are shown on the fin contour. (C) The contour is combinatorially sampled between every keypoint pair to produce a set of local, semi-local and global subsections. }
\label{fig:combinatorialSampling}       
\end{figure*}
\vspace{-35pt}
\section{Biometric Contour Encoding}
\label{sec:idEncoding}

In this section we develop a method of encoding smooth object shape suited to individual white shark fin representation. It  enables efficient and accurate individual recognition whilst being robust to noisy, partially occluded input generated by automatic shape extraction.

Global shape descriptions, as used in~\cite{stewman06}, maximise inter-class variance but are sensitive to partial occlusions and object-contour detection errors, while the removal of nuisance variables such as in- and out-of-plane rotation rely upon computing point correspondences and inefficient pairwise matching. 

By contrast, the semi-local descriptions of~\cite{arandjelovic11,arandjelovicO12} are robust and allow efficient matching, but their encoding of inter-class variance will always be sub-optimal. To maximise the descriptiveness of features, we utilise \textit{both} semi-local and global shape descriptions with a framework extending that used to generate fin candidates.  
\\ \ \\
 \textbf{Edge Refinement.} Our segmentation and contour partitioning framework so far produces  descriptions of  fin contours, but it does not resolve \textit{to sufficient resolution} the  fin shape along trailing edge and tip  vital to distinguishing individuals within shark populations \citep{anderson2011, Bonfil2005}. To recover this detailing we apply border matting in a narrow strip either side of region boundaries using the local learning method and code of \citet{zheng09}. This produces an opacity mask $\alpha$ which defines a soft segmentation of the image  $(\alpha_i \in [0,1])$. We obtain a binary assignment of pixels (by threshold $0.5$) to separate fin and background, and extract the resulting high resolution contour of best Chamfer distance fit as a precursor to biometric encoding. Full details of this edge refinement procedure can be found in~\cite{Hughes2015a}. 
\\ \ \\
\textbf{Generating Boundary Subsections.}
As a first step towards a biometric encoding, we detect salient boundary keypoints on the extracted contour strokes to produce stably recognisable contour subsections that serve as descriptor regions. For keypoint detection we use the same approach as that used for detecting keypoints when generating fin candidates, as described in Section~\ref{sec:strokeModel}. To generate boundary subsections, we resample fin candidates to a fixed resolution of 1024 pixels and compute $D(u,\sigma)$ in Equation~\ref{eq:dog}, re-parameterised with $\sigma=2$ and $m=8$. Subdivision by these keypoints yields ${50 \choose 2}=1225$ contour subsections\footnote{Taking as keypoints the $n=48+2$ largest local maxima of $D$, that is plus the start and end points of the contour, the putative fin boundary is sampled between every keypoint pair. }. Note that for reference images, we encode subsections in both directions. For test images, we encode in one direction only. As a result, later subsection matching does not need to consider the directions. The approach is illustrated in Figure~\ref{fig:combinatorialSampling}.
\ \\ \\
\textbf{Boundary Descriptors.} Following the generation of boundary subsections, the task is to encode their shape information. We investigate two regimes for subsection description: the standard DoG norm (DoG\textsubscript{N}) as defined in Equation~\ref{eq:dog}, and the  boundary descriptor of \cite{arandjelovicO12}. DoG\textsubscript{N} provides a number of properties relevant to biometric contour encoding: first, the associated metric is suitable for establishing similarity between descriptors, meaning contour sections can be matched efficiently. Secondly, by varying the parameters $\sigma$ and $m$, the description can be tuned to encode different components of shape scale-space. Third, the descriptor is rotation invariant and robust to  changes in viewpoint (see Figure~\ref{fig:matches}). 

We also consider the boundary descriptor of \cite{arandjelovicO12} composed of a vector of boundary normals, denoted $\mathcal{N}(u,\sigma)$. At each vertex the normal vector of the contour is computed and the two orthogonal components are concatenated to yield the  descriptor:\vspace{-6pt}
\begin{equation}
\label{eq:normalDescriptor}
\mathcal{N}(u,\sigma)=(G(u,\sigma)*x(u),G(u,\sigma)*y(u))
\end{equation} 
This normal descriptor lacks rotational invariance. This is overcome by aligning the  ends of each subsection with a fixed axis as a precursor to descriptor computation. 

\begin{figure}[hb]
\centering
\includegraphics[width=0.45\textwidth]{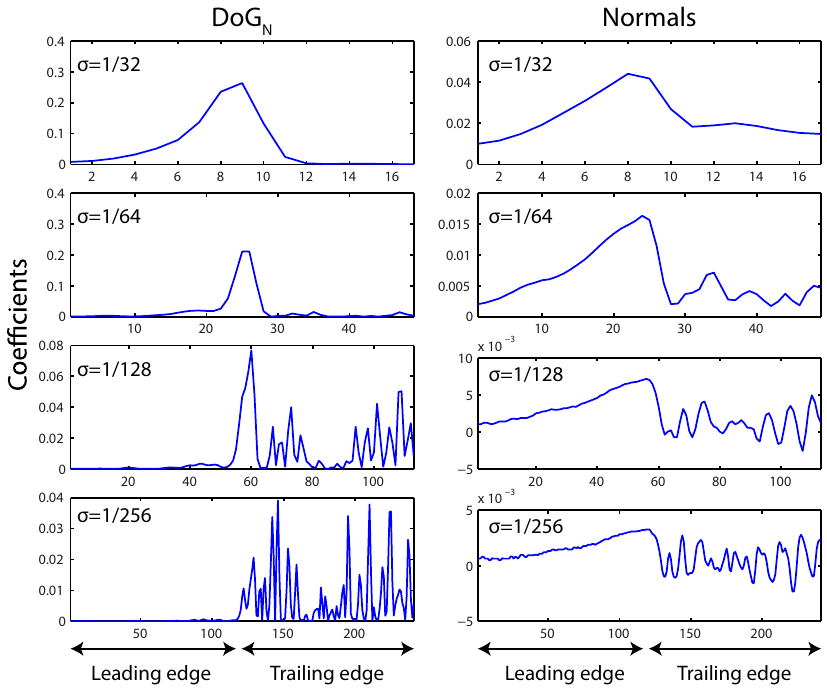}
\caption{DESCRIPTORS FOR ENCODING INDIVIDUAL FIN SHAPE: We utilise the DoG\textsubscript{N} and Arandjelovic's normal descriptor as a feature pool for characterising individuality. It can be seen that both location on the segment~(x-axis) and scale-space band ($\sigma$) are encoded by the descriptors.}
\label{fig:coeffs}       
\end{figure}

As  illustrated in Figure~\ref{fig:coeffs} over the entire fin segment, both DoG\textsubscript{N} and Arandjelovic's normal descriptor provide spatial and scale selectivity. 

\section{  Identification   Baseline via LNBNN}
\label{sec:LNBNN}
As noted by \citet{boiman08}, information is lost in processes such as vector quantisation. For this reason, we utilise a scoring mechanism inspired by the local naive Bayes nearest neighbour (LNBNN) classification algorithm \citep{mccann12}, and similar to that employed by \citet{crall13} in the context of patterned species individual identification, to provide a recognition baseline. 

\begin{figure*}[ht]
\centering
\includegraphics[width=0.8\textwidth]{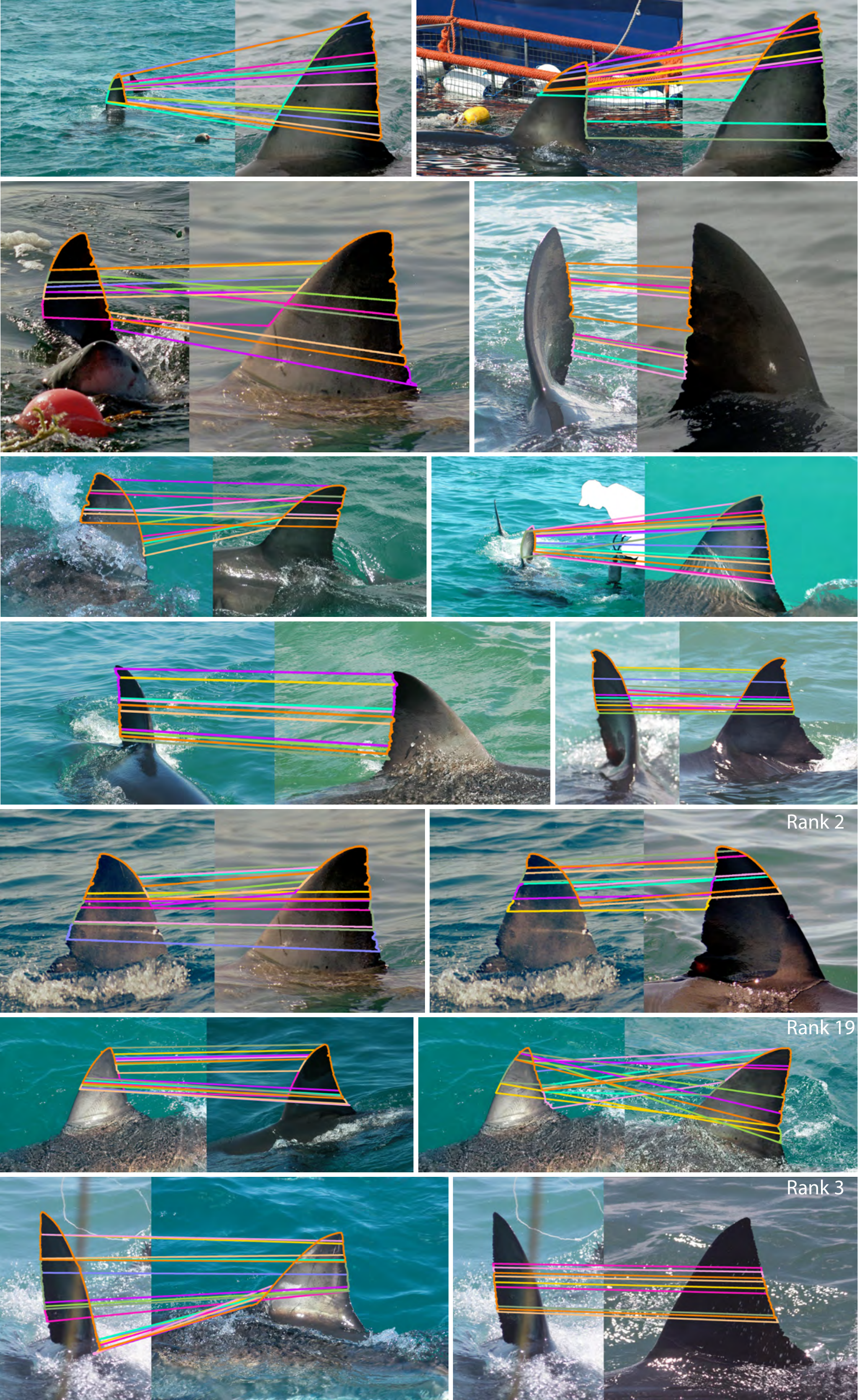}
\caption{LNBNN INDIVIDUAL IDENTIFICATION EXAMPLES: left images are queries and right ones are predicted individuals. Coloured lines indicate start and end of the ten sections contributing most evidence for the matched individual. For illustration of false matches, bottom three rows, left pairs, show misidentifications while correct matches are shown right. All example matches are obtained using multiscale DoG\textsubscript{N} descriptors combined using the LNBNN classifier. Out of respect for their privacy, the human subject appearing in row 3, column 2, was masked out of the image prior to publication, but only after fin detection and photo-identification results had been obtained.}
\label{fig:matches}       
\end{figure*}

Specifically, denoting the set of descriptors for a query object $D_Q$, for each query descriptor $d_i \in D_Q$ and for each class $c\in C$, we find two nearest neighbours ($NN_c(d_i),NN_{\bar{C}}(d_i)$) where $\bar{C}$ is the set of all classes other than $c$. Using the shorthand $\delta(NN_{\cdot}) = ||d_i - NN_{\cdot}(d_i)||^2$,
queries are classified according to: \vspace{-1pt}
\begin{equation}
\label{eq:nbnn1}
\hat{C}=\operatorname*{arg\,max}_C \sum^{|D_Q|}_{i=1} f(d_i,c)
\end{equation}\vspace{-10pt}
\begin{equation}
\label{eq:localScore}
f(d,c)=
\begin{cases}
\delta(NN_{\bar{C}}) - \delta(NN_c) & \delta(NN_{\bar{C}})>\delta(NN_c) \\
0 & \text{otherwise} \\ 
\end{cases}
\end{equation}
This decision rule can be extended to a multi-scale case. Letting $S=\{\sigma_1,...,\sigma_j,...,\sigma_v\}$ denote the set of scales for which we compute descriptors, the multi-scale decision rule linearly combines the contribution of the descriptors at each scale (see also top of Figure~\ref{fig:overview3}):\vspace{-5pt}
\begin{equation}
\label{eq:nbnn2}
\hat{C}=\operatorname*{arg\,max}_C\displaystyle\sum_{j=1}^{v}w_j\cdot \displaystyle\sum_{i=1}^{|D_Q^j|}f(d_i^j,c)
\end{equation}
\\ 
\textbf{Implementation Details.} To achieve scale normalisation, each contour subsection is re-sampled to a fixed length of $256$ pixels. DoG\textsubscript{N} and normal descriptors are computed at filter scales $S=\{1,2,4,8\}$, with a constant value of $m=2$ in the DoG\textsubscript{N} case. Each descriptor is $L2$ normalised to allow similarities between descriptors to be computed using Euclidean distance. FLANN \citep{muja09} is employed to store descriptors and to perform efficient approximate nearest neighbour searches. Classification is performed at each scale separately for both descriptor types and then combined, with each  scale weighted equally~($w_j=1$). \vspace{-12pt}

\ \\ \\
\textbf{Dataset.} In order to benchmark individual fin classification, we use a dataset representing $85$ individuals and consisting of $2456$ images (see Acknowledgements for data source).  For each individual there are on average $29$ images (standard deviation of $28$). The minimum number for an individual was two. As such, when the dataset was split into labelled and test images, just one labelled training example was selected to represent each shark. The remaining $2371$ images were used as queries all of which show at least $25\%$ of  the fin's trailing edge. They exhibited significant variability in waterline and white splash occlusion, viewpoint, orientation and scale (see Figure~\ref{fig:teaser} and Figure~\ref{fig:matches} for example images).
\\ \ \\
\textbf{Performance Evaluation Measures.} Two measures are reported for performance evaluation. Both are based on average precision as the classifier returns a ranked list of candidate identities, each associated with a score as computed according to Equations~\ref{eq:nbnn1} or~\ref{eq:nbnn2}. The first is AP, computed for all test images.  For the second, we compute AP for each individual and then take the mean of the individual AP scores~(mAP). This second measure avoids bias towards individuals with large numbers of test images. In each case, AP is computed as area under precision-recall curves computed directly using the individuals' scores, in contrast say to the ranking employed in~\cite{everingham14}.
\\ \ \\
\textbf{Results.}
The mAP and AP scores for DoG\textsubscript{N} and normal-based individual identification are shown in Table~\ref{tab:idResults}. Overall, our contour stroke model for fin detection combined with a combinatorial biometric contour encoding proves suitable for the task of individual fin identification. For DoG\textsubscript{N}, as reported in~\cite{Hughes2015} for one-shot-learning, of the $2371$ query instances presented to the system,  a particular shark is correctly identified with a mAP of~$0.79$.
\begin{table}[t]
\caption{INDIVIDUAL LNBNN ID RESULTS }
\label{tab:idResults}       
\begin{tabular}{llllll}
\hline\noalign{\smallskip}
\multicolumn{6}{c}{1 training image per class (1-shot-learning): 2371 queries}  \\
\noalign{\smallskip}\hline\noalign{\smallskip}
Encoding & $\sigma=8$ & $\sigma=4$ & $\sigma=2$ & $\sigma=1$ & combined\\
\noalign{\smallskip}\hline\noalign{\smallskip}
AP:DoG\textsubscript{N} & 0.63  & 0.72  & 0.69  & 0.49  & 0.76 \\
AP:Norm & 0.33  & 0.70  & 0.72  & 0.65  & 0.72 \\
mAP:DoG\textsubscript{N} & 0.67 & 0.74 & 0.73  & 0.56& 0.79 \\
mAP:Norm & 0.49 & 0.75 & 0.76 & 0.73 & 0.76 \\
\noalign{\smallskip}\hline\noalign{\smallskip}
\multicolumn{6}{c}{2 training images per class: 2286 queries }  \\
\noalign{\smallskip}\hline\noalign{\smallskip}
\multicolumn{3}{l}{AP:SIFT}  &  &  & 0.20 \\
\multicolumn{3}{l}{mAP:SIFT}  &  &  & 0.35 \\
AP:DoG\textsubscript{N} & & & & & 0.81 \\
mAP:DoG\textsubscript{N} & & & & & 0.83 \\
\noalign{\smallskip}\hline
\end{tabular}
\end{table}
Figure~\ref{fig:matches} illustrates such examples of fin matches. 
An examination of recognition performance for high quality fin detections ($F^g\textsubscript{inst}$$>0.9$) provides insight into the effect of fin detection on individual identification. Of $217$ such detections, where additionally, the entire fin contour was clearly visible, $82\%$ were correctly identified with a mAP of $0.84$. In $91\%$ of cases, the correct identity was returned in the top ten ranks. Thus, approximately $9\%$ of fin instances could not be classified correctly, independent of the quality of the detected contour.

The results  demonstrate the benefit of combining DoG\textsubscript{N} descriptors computed for independent scale-space components of fin shape, as shown by a $6.7\%$ gain in AP performance from AP=$0.72$ to AP=$0.76$ compared to that obtained using any individual scale alone. 

The  normal encoding also proves suitable for individual recognition, with AP of 0.72 and mAP of 0.76, although the best performance obtained with this descriptor type falls below the multi-scale DoG\textsubscript{N} approach. 

\begin{figure}[hb]
\centering 
\includegraphics[width=0.4\textwidth]{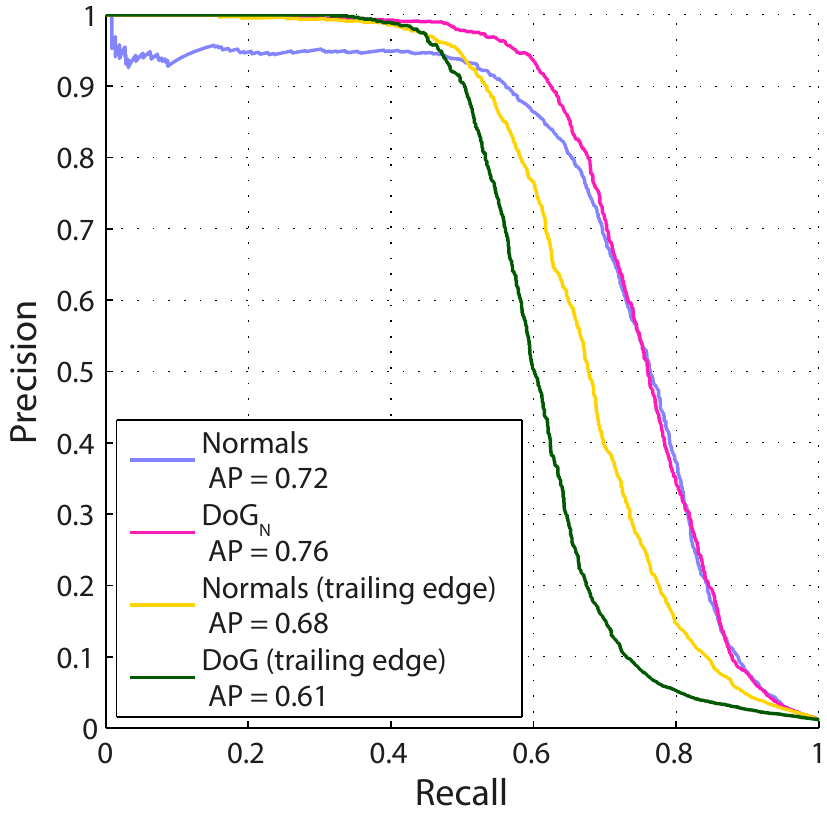}
\caption[Comparing performance for different encoding types]{PRECISION-RECALL CURVES FOR LNBNN. Precision-recall curves for DoG\textsubscript{N} and normal fin encodings, comparing identification via whole fins and just trailing edges.}
\label{fig:PRNormVsDoG}
\end{figure}

Figure~\ref{fig:PRNormVsDoG} shows precision-recall curves for DoG\textsubscript{N} and normal encoding types. It can be seen that the recognition performance difference between the two feature types occurs in the high precision region, with a normal encoding providing recognition precision of less than one for almost all values of recall. When descriptors corresponding to the trailing edge of fins alone are considered, the normal encoding provides superior recognition to that obtained using DoG\textsubscript{N}, but nevertheless remains inferior to that obtained using a multi-scale DoG\textsubscript{N} representation of the whole fin. 

\vspace{-8pt}

Finally, we observe that the DoG\textsubscript{N} and normal approaches produce different predictions on a significant set of samples, pointing towards an opportunity in combining  these classifiers, depending on fin structure. This complementarity is exploited in Section~\ref{sec:finspace}.

\newpage
\noindent \textbf{Comparison with Off-the-shelf Features.}
To put the performance of our biometric contour representation in context, we report individual fin identification results using a methodology previously applied to patterned species individual recognition \citep{crall13}. In our case, a sparse, affine covariant SIFT encoding \citep{mikolajczyk04} of fin shape and surface texture is generated by detecting features centred within closed regions, created by drawing straight lines between the two ends of detected fin stokes (illustrated using dashed lines in Figure~\ref{fig:teaser2}). As before, LNBNN (Equations~\ref{eq:nbnn1} and~\ref{eq:localScore}) is used for individual classification. In this experiment (and only this experiment) two training images are used per individual, one for each side of the fin, leaving 2286 query images. 

\begin{figure}[t]
\centering
\includegraphics[width=84mm]{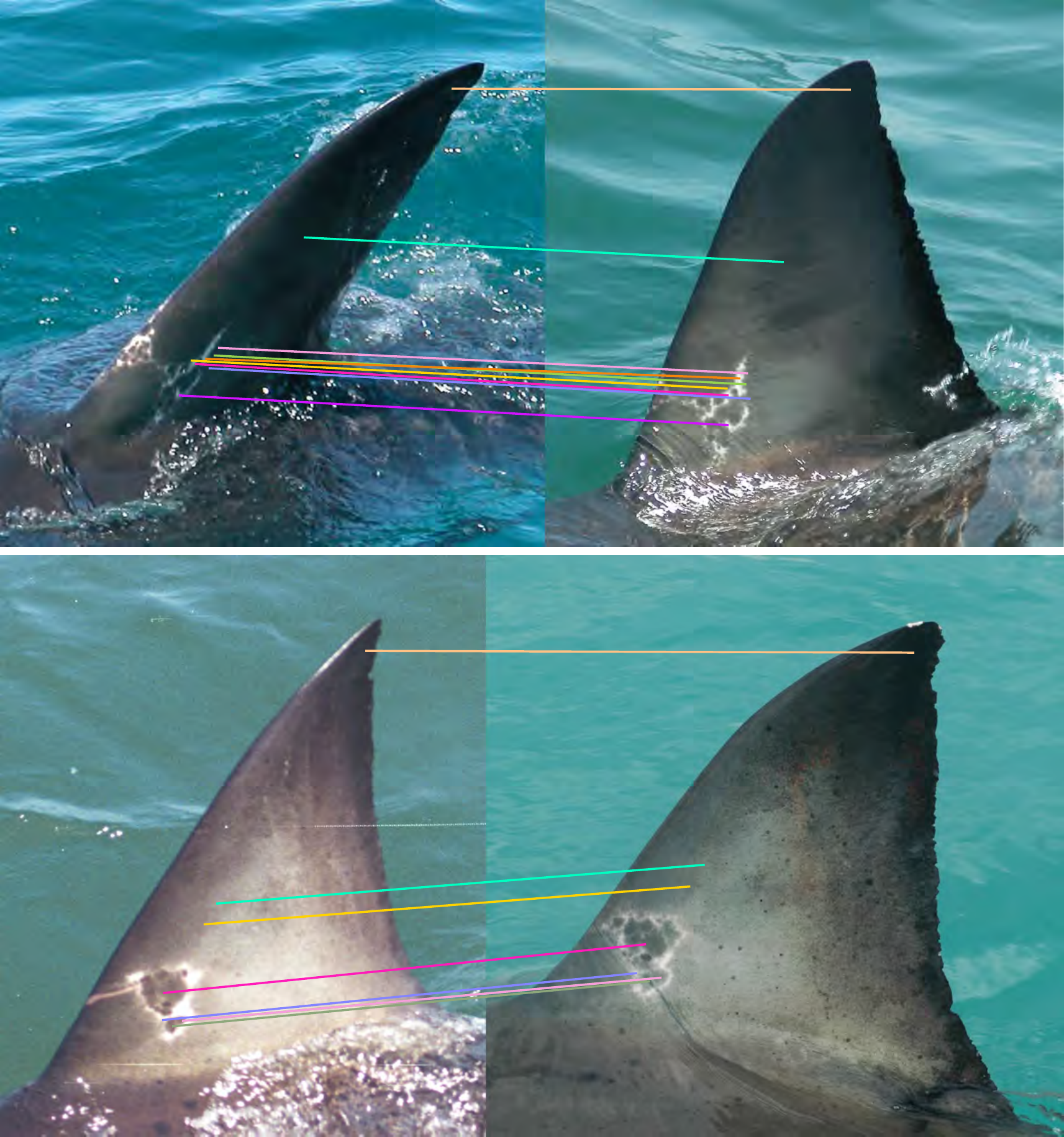}
\caption[Comparing performance for different encoding types]{EXAMPLE IDENTIFICATIONS USING AFFINE-COVARIANT SIFT DESCRIPTIONS: Rarely, fins misclassified using biometric contour representations are correctly identified using surface texture descriptors. Here, two such examples are shown, with query images on the left of each pair. The coloured lines represent discriminative feature matches (as evaluated by the value of $f(d,c)$ in Equation~\ref{eq:localScore})\vspace{-15pt}}
\label{fig:siftMatches}
\end{figure}

Results in Table~\ref{tab:idResults} unambiguously demonstrate the superiority of our biometric contour representation over one describing surface texture, for individual fin identification. Using SIFT features, fins are identified with mAP of 0.35 (AP=0.2). Using exactly the same training data, this compares with mAP of 0.83 using the combinatorial multi-scale DoG\textsubscript{N} encoding (AP=0.81).  Interestingly however, 45 fin instances, misclassified using biometric contour encoding, are correctly identified using SIFT, with examples shown in Figure~\ref{fig:siftMatches}. Noting that the permanence of fin surface markings additionally captured by 2D features such as SIFT is disputed \citep{robbins13}, this observation nevertheless suggests that texture-based representations may have potential utility, at least for a sub-set of the population and over short observation windows.

\vspace{-12pt}
\section{Construction of a Population-wide Fin Space }
\label{sec:finspace}
\begin{figure}[b]
\centering
\includegraphics[width=0.45\textwidth]{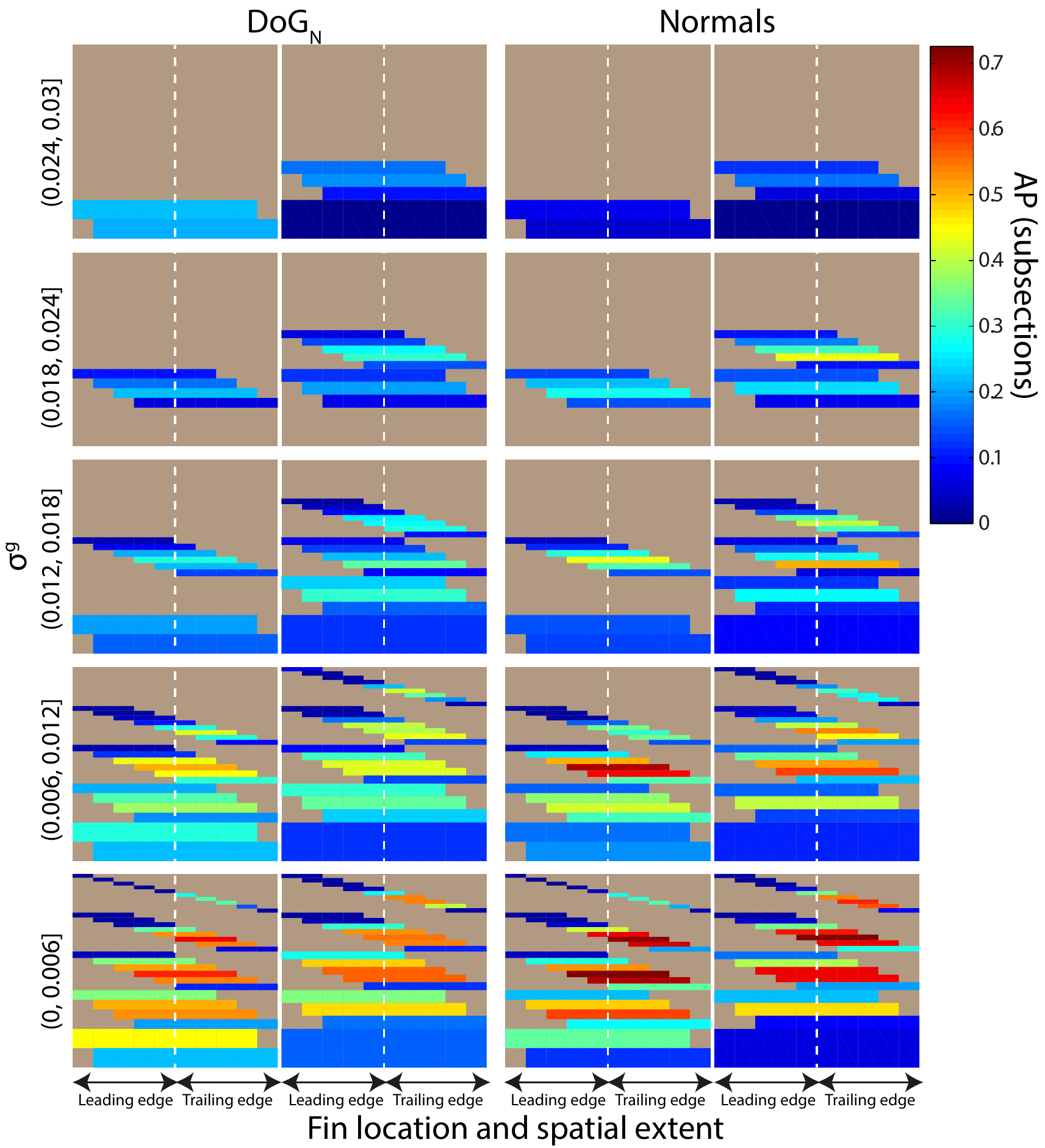}
\caption{FIN SPACE AND  LOCALISATION OF INDIVIDUALITY. Organising visual descriptors indexed over spatial location (x-axes) and extent  on the fin (dotted line marks fin tip), and filter scale (rows) allows for the learning of population-wide distinctiveness properties associated with the anatomic fin locations. Colouration depicts the distinctiveness of bins with respect to animal individuality, as quantified by classification AP at the subsection level.}
\label{fig:locLenFreq}       
\end{figure}

\vspace{-3pt}
In this section, we introduce a globally normalised cross-class~(cross-individual) coordinate system over \textit{both} descriptors DoG$_N$ and normals, i.e. a global `fin  space', in which we embed fin descriptors along the dimensions of descriptor type, spatial location and spatial extent on the fin contour, as well as along feature scale. The resulting $4D$ fin space is illustrated in Figure~\ref{fig:locLenFreq}.

This space   allows   for reasoning about and learning of population-wide properties using anatomically interpretable dimensions; be that  to 1) quantify the distinctiveness of feature descriptors by their type, location or extent on the fin, or to 2) move from a non-parametric and linear method of cue combination to one that non-linearly learns how to combine indexed evidence from across the fin space. 
Importantly, this entails learning a single model for the species, one which can be seen as characterising a species-specific pattern of individual distinctiveness, and not one that learns a pattern of uniqueness solely for any given individual.
\\ \ \\
\textbf{Embedding  Descriptors into Fin Space.} The fabric of  the proposed fin space can   be described as subdividing the leading and trailing edges of fins into ($n=5$) equally sized partitions\footnote{As the lengths of either edge of the fin are not necessarily the same, the size of the partitions on the leading edge are not necessarily the same as those on the trailing edge.}. We then consider  every connected combination of partitions yielding $55 $ spatial bins for each of the two feature types. 

\begin{figure}[h]
\centering
\includegraphics[width=0.47\textwidth]{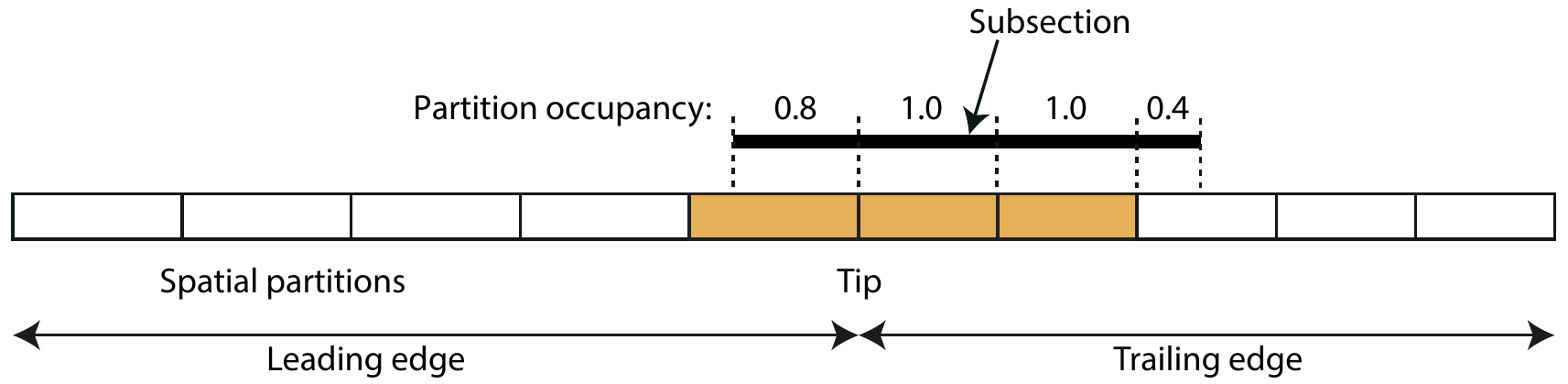}
\caption{SPATIAL EMBEDDING OF FIN PATTERNS. Example of a subsection mapped to a spatial bin (shown in yellow) covering 3 partitions.\vspace{-20pt} }
\label{fig:spatialMapping}       
\end{figure}

As illustrated in Figure~\ref{fig:spatialMapping}, fin subsections can be mapped to spatial bins by first assigning them to partitions - a subsection is said to occupy a partition if it occupies more than half of it. Finally, each subsection is assigned to the spatial bin that corresponds to the set of partitions it occupies. Scale-space partitioning is achieved by dividing filter scale into five bins.  
\vspace{-8pt}
\\ \ \\
More formally, this yields an overall set of  bins given by $B=\{(\sigma_1^g,\sigma_2^g],...,(\sigma_k^g,\sigma_{k+1}^g],...,(\sigma_{m-1}^g,\sigma_m^g]\}$ and the set of filter scales is $S^g = \{\sigma_1^g,...,\sigma_k^g,...,\sigma_m^g\}$. Here  $g$  denotes that filter scale is considered as a proportion of the fin contour length~globally. 

Defined globally, the filter scale of the $i^{th}$ subsection descriptor computed at scale $j$ (as in Equation~\ref{eq:nbnn2}) can be expressed as $\sigma^g_{i,j} =\nicefrac{\sigma_j}{l_n} \cdot p$ where $p$ expresses the length of the subsection as a proportion of the length of the fin contour, and $l_n$ is the number of samples used to encode the subsection. Having computed $\sigma^g_{i,j}$, the descriptor is mapped to the corresponding bin. 
\vspace{-16pt}
\section{Non-Linear Model Exploiting Fin Space}
\vspace{-8pt}
\label{sec:ID}
In this section we show that learning   distributions of reliable match locations in fin space can significantly improve  identification rates compared to the baseline. This appreciates the fact that certain feature \textit{combinations} in fin space are common and not individually discriminative in sharks, whilst others are highly distinctive. To implement a practical approach that  captures such species-specific information, we learn a non-linear map from  patterns of  matching locations    in fin space to  likelihoods of reliability for identification.  
\\ \ \\
\textbf{Obtaining Scoring Vectors from Fin Space.} As in the baseline case, for each query descriptor (representing the input image) and for each class (representing the individuals),  we find the nearest reference descriptor in that class, i.e. perform max-pooling over the class. As described in Section \ref{sec:LNBNN}, based on the distance to that nearest neighbour and the distance to the nearest neighbour in \textit{another} class, we compute a local match score according to Equation~\ref{eq:localScore}. 

\begin{figure}[t]
\centering
\includegraphics[width=0.49\textwidth]{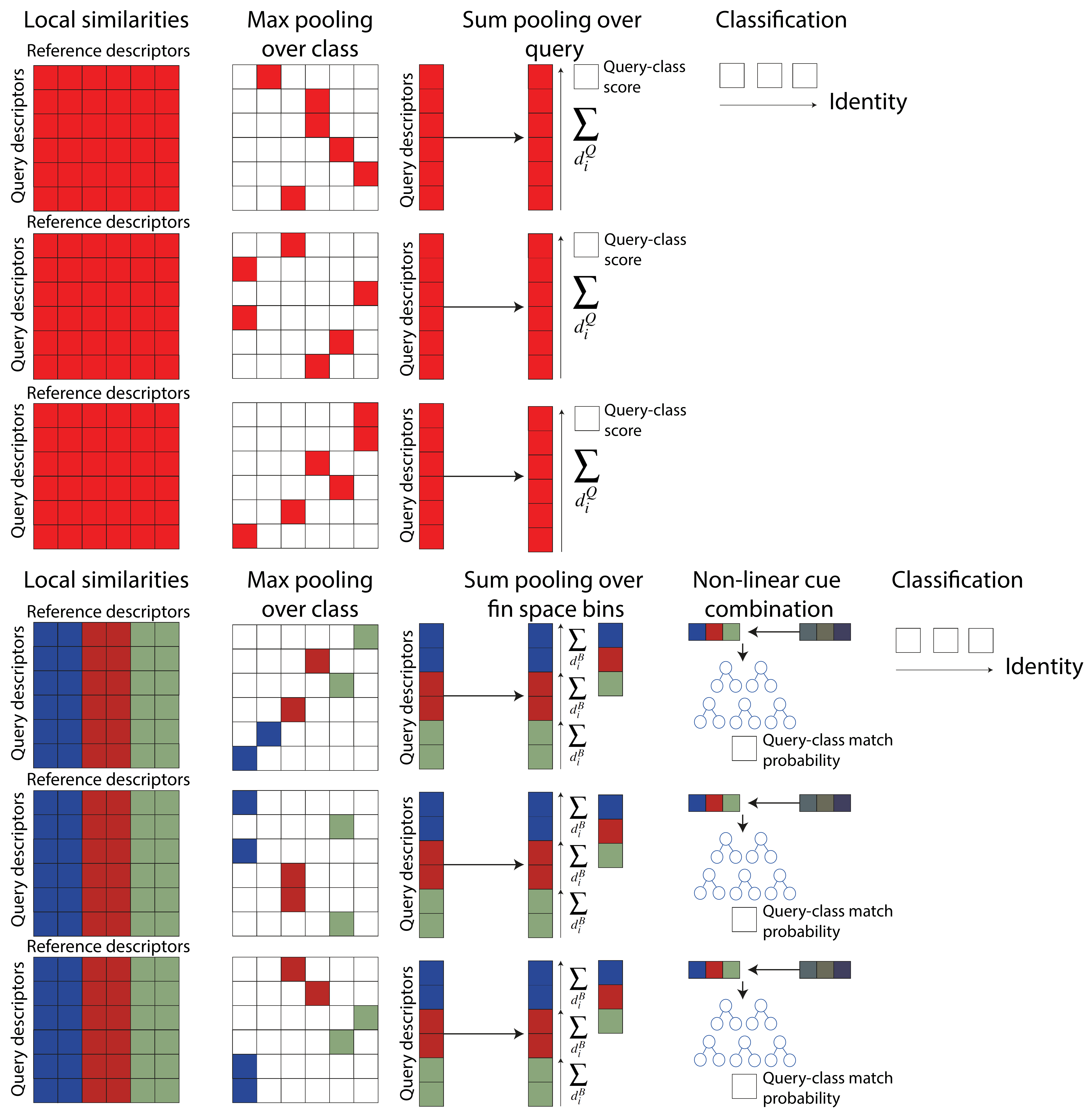}
\caption{COMPARISON OF BASELINE (top) AND FIN-SPACE IDENTIFICATION SCHEME (bottom). The two paradigms are illustrated proceeding from left to right. By associating descriptor matching scores~(left column) to reference locations in a global fin space (colouration), the improved scheme (bottom)  accumulates information not into a single, class-specific scalar (top approach), but forms a scoring vector that encodes the pattern of matchings over fin space. Identity is then judged via a random forest based on the learned reliability of the matching patterns.
}
\label{fig:overview3}       
\end{figure}

Now, instead of sum-pooling local match scores over class labels directly, as performed in Equations~\ref{eq:nbnn1} and~\ref{eq:nbnn2}, we first project local match scores into fin space via their associated reference descriptors, and  then perform sum-pooling over fin space bins (see Figure~\ref{fig:overview3}). As a result, for each class and for each discrete fin space location, we obtain a score. These scores form a vector of dimensionality equal to the cardinality of fin space. As such, each query-class comparison yields such a vector.\\ \ \\
\textbf{Learning a Non-Linear Identification Model.}
The described procedure rearranges matching information so that the scoring pattern as observed spatially and in scale-space along the fin, as well as over descriptor types, is made explicit by the scoring vector. We now analyse the structure of scoring vectors over an entire population of fins to learn and predict their reliability for inferring animal identity.
This procedure is designed to selectively combine descriptor evidence (see Section \ref{sec:LNBNN}), exploit the observed variance in local distinctiveness (see Figure~\ref{fig:locLenFreq}),
and
address potential correlations between features in fin space. 
To allow for complex, non-linear relationships between scoring structures and identification reliability, we select a 
random forest classifier to implement the mapping.

\begin{figure}[hb]
\centering \vspace{-12pt}
\includegraphics[width=0.4\textwidth]{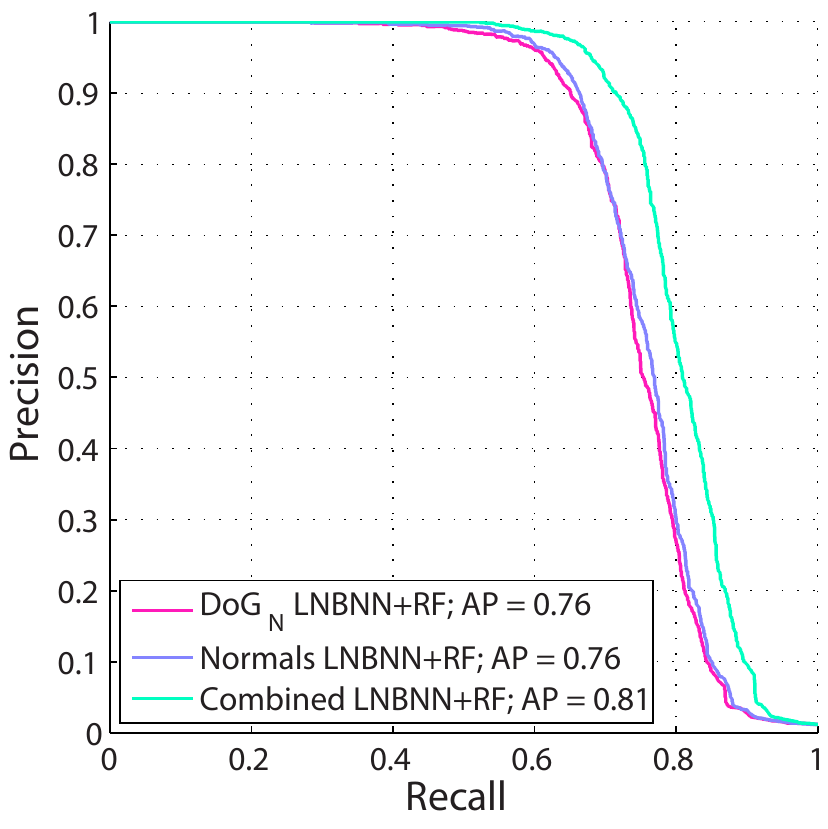}
\vspace{-8pt}
\caption{RESULTS OF IDENTIFICATION USING THE FIN SPACE APPROACH. Precision-recall curves reported considering each of the descriptor types separately (effectively training the random forest on only half the fin space), as well as considering the full dual descriptor set.  }
\vspace{-10pt}
\label{fig:PR_RF}       
\end{figure}

Practically, we train the random forest to map from query-class scoring vectors to probability distributions over binary match category labels `query-is-same-class' and `query-is-not-same-class'. Importantly, performing two-fold cross-validation, the dataset is split randomly by individual, and not by query, when training and evaluating the classifier. This ensures that what is learned generalises across the species and does not over-fit the individuals in the present dataset.
\\ \ \\
\textbf{Final Results.} Evaluation is performed by reporting AP and precision-recall curves over the same 2371 queries as used to obtain the identification baselines in Section~\ref{sec:LNBNN}. We present the results in Figure~\ref{fig:PR_RF}.
It can be seen that, overall, the final fin space approach achieves an AP of $0.81$, representing 7\% and 12\% performance gains over the DoG$_N$ and normal baselines, respectively. The results also clearly demonstrate the benefit of selectively combining both descriptor types -  precision measures are improved or kept across  the entire recall spectrum for a combined, dual descriptor approach. 


\section{Conclusions and Future Work}
\vspace{-10pt}
\label{sec:conclusions}
We have described a vision framework for automatically identifying individual great white
sharks as they appear in unconstrained imagery as used by white shark researchers.  To do so, we have first  described a
contour stroke model that partitions ultrametric contour maps and detects fin objects based on
the resulting open contour descriptions.   We have shown that this process simultaneously
generates fin object candidates and separates them from background clutter. 

Secondly, a multi-scale
and combinatorial method for encoding smooth object boundaries biometrically  has been described. In combination with an LNBNN classifier, the method is both discriminative and robust, and shows  individual shark
fin identification performance at a level of AP=$0.76$ when employed using a multi-scale DoG descriptor in a one shot learning paradigm. 

Thirdly, we have introduced a domain-specific `fin space' which indexes fin shapes spatially, by filter scale and along descriptor types. We have measured  the distinctiveness for individual shark identification of different regions in this space, providing some insight into the distribution of individuality over the fin. 

Finally, we have proposed a shark fin identification framework that achieves an AP=$0.81$ outperforming the baseline system published in~\cite{Hughes2015}.  In essence,  we achieved this improvement by introducing
a non-linear recognition model, which integrates different descriptors and operates based on a population-wide, learned model for predicting identification reliability from matching  patterns in fin space.

For the species at hand, we conclude practical applicability
at accuracy levels ready to assist human identification efforts without a need for any manual
labelling. The approach may therefore be  integrated to enhance large scale citizen science \citep{zoo2011,ibe2015,Duyck2014} for ecological data collection of white sharks. A related project to make available this work to the biological research community is  underway~\citep{SoS2016}.

Furthermore, we expect our framework to generalise to other classes of smooth biometric entity,
in particular marine life exhibiting individually distinctive fin and fluke contours such as various other  species of shark and whale, e.g. humpback whales \citep{ranguelova04}. 

\vspace{-18pt} 
\section*{Dataset}\vspace{-8pt}
The dataset "FinsScholl2456" containing 2456 images of great white
sharks and their IDs was used in this paper. Since the authors and host institution hold no copyright, to obtain a copy please directly contact: \\ Michael C. Scholl,
Save Our Seas Foundation (CEO),
Rue Philippe Plantamour 20, CH-1201,
Geneva, Switzerland; \ Email:
Michael@SaveOurSeas.com

\clearpage
\vspace{-8pt}
\section*{Acknowledgements}\vspace{-12pt}
B.H. was supported by EPSRC grant EP/E501214/1.  We gratefully acknowledge Michael
C. Scholl and the Save Our Seas Foundation for allowing the use of fin images and ground truth labels.\vspace{-8pt}


\bibliographystyle{spbasic}      
\bibliography{egbib}   

%
%

\end{document}